\documentclass[a4paper,fleqn]{cas-dc}

\usepackage[authoryear,longnamesfirst]{natbib}
\usepackage{amssymb}
\usepackage{subfigure}
\usepackage{hyperref}
\usepackage{listings}
\usepackage{xcolor}
\usepackage{setspace}
\usepackage[lined,boxed,commentsnumbered,procnumbered,ruled]{algorithm2e}
\usepackage{amsmath}
\usepackage{algpseudocode}  
\usepackage[authoryear,longnamesfirst]{natbib}

\def\tsc#1{\csdef{#1}{\textsc{\lowercase{#1}}\xspace}}
\tsc{WGM}
\tsc{QE}
\begin{document}
\let\WriteBookmarks\relax
\def\floatpagepagefraction{1}
\def\textpagefraction{.001}

\shorttitle{\textbf{TONE}: A $3$-\textbf{T}iered \textbf{ON}tology for \textbf{E}motion analysis}    

\shortauthors{Srishti Gupta, Piyush Kumar Garg, Sourav Kumar Dandapat}  

\title [mode = title]{\textbf{TONE}: A $3$-\textbf{T}iered \textbf{ON}tology for \textbf{E}motion analysis}  

% \tnotemark[<tnote number>] 



\author[1]{Srishti Gupta}[orcid=0000-0003-1043-1296]

\cormark[1]
\fnmark[1]
\ead{srishti_2021cs38@iitp.ac.in}
% \ead[url]{<URL>}
\credit{Conceptualization, Methodology, Formal Analysis, Writing - Original Draft}

\affiliation[1]{organization={Computer Science and Engineering, Indian Institute of Technology Patna},
            addressline={Bihta}, 
            city={Patna},
%          citysep={}, % Uncomment if no comma needed between city and postcode
            postcode={801103}, 
            state={Bihar},
            country={India}}

\author[1]{Piyush Kumar Garg}[orcid=0000-0003-2266-9605]

% \cormark[1]
\fnmark[1]
\ead{piyush_2021cs05@iitp.ac.in}
% \ead[url]{<URL>}
\credit{Conceptualization, Methodology, Software}

\author[1]{Sourav Kumar Dandapat}[orcid=0000-0003-2043-2356]
% \cormark[3]
% \fnmark[3]
\ead{sourav@iitp.ac.in}
% \ead[url]{<URL>}
\credit{Writing - Review \& Editing, Supervision}

% \cortext[cor1]{Corresponding author}

\fntext[1]{Both authors contributed equally to this work and are jointly the first authors.}

\begin{abstract}
Emotions have played an important part in many sectors, including psychology, medicine, mental health, computer science, and so on, and categorizing them has proven extremely useful in separating one emotion from another. Emotions can be classified using the following two methods: (1) The \textit{supervised} method's efficiency is strongly dependent on the size and domain of the data collected. A categorization established using relevant data from one domain may not work well in another. (2) An \textit{unsupervised} method that uses either domain expertise or a knowledge base of emotion types already exists. Though this second approach provides a suitable and generic categorization of emotions and is cost-effective, the literature doesn't possess a publicly available knowledge base that can be directly applied to any emotion categorization-related task. This pushes us to create a knowledge base that can be used for emotion classification across domains, and \textit{ontology} is often used for this purpose. In this study, we provide \textbf{TONE}, an emotion-based ontology that effectively creates an emotional hierarchy based on Dr. Gerrod Parrot's group of emotions. In addition to ontology development, we introduce a semi-automated vocabulary construction process to generate a detailed collection of terms for emotions at each tier of the hierarchy. We also demonstrate automated methods for establishing three sorts of dependencies in order to develop linkages between different emotions. Our human and automatic evaluation results show the ontology's quality. Furthermore, we describe three distinct use cases that demonstrate the applicability of our ontology.

\end{abstract}

% \begin{highlights}
% \item 
% \item 
% \item 
% \end{highlights}

% Keywords

\maketitle

\section{Introduction} \label{introduction}
\textbf{Emotion categorization}, a contentious topic in various fields, is the process of setting apart or distinguishing one emotion from another. Numerous domains, like psychology, medicine, history, computer science, etc., might benefit from categorizing emotions for their tasks. In the field of NLP, with the right emotion categories, activities like response creation, product review analysis, etc. enhance with improved emotion detection. This categorization can be achieved by using either \textit{supervised} or \textit{unsupervised} methodology. In order to obtain a categorization using a supervised method, enormous amounts of data must be collected, annotated, and after analysis, one could develop a categorization. However, the suitability of it is restricted to that particular domain pertaining to the data. Therefore, in order to achieve accurate categorization for a different domain, it is necessary to replicate the entire data-gathering procedure. On the contrary, the generation of such categorization with the usage of an unsupervised approach is based on expert knowledge in the domain. This process would assist in building a general knowledge base that properly categorizes emotions and maps their interactions for use in diverse applications. Usually, an \textbf{ontology} (\cite{narayanasamy2021ontology}, \cite{article1}, \cite{article3}, \cite{garg2023ontodsumm} etc.) plays a significant role in achieving a systematic categorization that serves as a knowledge base. However, a major impediment in the literature is the lack of any such knowledge base, and the available emotion-based ontologies are extremely task-specific and not generic. In our work, we focus on emotion-based ontologies and delve deeper into their pitfalls. Prior works build an emotion-based ontology, which is either \textit{task-specific} or uses an existing \textit{models of emotion groupings} to build an ontology. The \textit{task-specific} emotion-based ontologies have been manually designed as in \cite{10.1007/978-3-642-24958-7_11}, \cite{article4}, and \cite{graterol2021emotion} to perform the machine-learning-related task called emotion classification. These existing emotion ontologies face the challenge of not describing a broad division of emotions for usage in other applications. Their constructed ontology neither provides a generalized division of emotions nor a vocabulary associated with the emotions. 

The other approach to building an emotion ontology is by using existing \textit{models of emotion groupings}. For instance, let's consider \cite{tabassum2016emotion}, which utilizes the well-renowned Plutchik's Wheel of Emotions \cite{plutchik1991emotions} to build an ontology called \textit{EmotiON}. They use the wheel of emotions to manually create the tree-structured representation of an ontology for the classification of emotions. They manually construct the relationships within the ontology, which makes it highly prone to inaccuracy. Additionally, the presented ontology \textit{can't be used directly for the classification task} due to the depiction of emotions in a non-hierarchical manner that does not offer the liberty to utilize emotions according to the granularity of the task, leading to employing all emotions at once. Moreover, if one wishes to create an ontology by considering any other grouping of emotions like \cite{Ekman1992-EKMATB}, \cite{SLazarus1994-SLAPAR-2} or \cite{cowen2017self} (details of these groupings are mentioned in Section~\ref{emotion}), they also propose a list of more than $10$ basic emotions which leads to a \textit{large number of nodes} at the first level in the hierarchy of the ontology. 
% Too many classes would mean an appropriate collection of data that clearly differentiates each emotion. Otherwise, with little data, it would lead to a case where there wouldn't be enough examples for a model to learn each class. 
Hence, an absence of an efficiently designed generic emotion-based ontology in the existing literature demands the introduction of an emotion ontology that appropriately classifies emotions, links them effectively, and consists of an extensive vocabulary for each emotion. 

To overcome the above-mentioned shortcomings, we propose an ontology, \textbf{TONE}: A $3$-\textbf{T}iered \textbf{ON}tology for \textbf{E}motion analysis, that uses the grouping of emotions proposed by Dr. Gerrod Parrott \cite{shaver1987emotion}. In TONE, we choose \textbf{Parrott's group of emotions} because it defines the classification of emotion in a tree-structured manner which could solve the problem of using emotions in a granular manner. Each level offers a broader classification of emotions and hence based on the required granularity, one could utilize TONE. For example, when considering an ontology built on Berkeley's grouping of emotions, it provides us with $27$ categories that could be computationally large in tasks like emotion classification. This hierarchical representation of emotions offered by Dr. Gerrod Parrott also perfectly suits the creation criteria of an ontology. So, in order to adapt this grouping into building an ontology, \cite{WU201287} define detailed procedures to be used that start with a root and then branch out into other classes, which is greatly aided by Parrott's group of emotions. Hence, our proposed ontology, TONE, considers Parrott's group of emotions because of its $3$-tiered tree-like classification of emotions. Furthermore, we offer an automated method for generating three types of dependencies among distinct emotions to account for unintended lapses while manually constructing relations. We wish to reduce the human effort needed to link various emotions. Lastly, a few works like \cite{doi:10.1177/0539018405058216}, Dr. Tom Drummond's vocabulary\footnote{\url{https://tomdrummond.com/leading-and-caring-for-children/emotion-vocabulary/}}, etc. focused on providing a vocabulary associated with a few emotions. For example, Dr. Tom Drummond came up with a vocabulary of emotions by collecting words from Thesaurus to educate the children regarding emotions. However, these \textit{vocabularies are rather limited}. Hence, we also introduce a semi-automated method of creating an extensive vocabulary for each emotion in our ontology.  
% According to them, one must begin with constructing an upper-level conceptual framework followed by creating the low-level class hierarchy. Whereas the emotions proposed by Dr. Gerrod Parrott \cite{shaver1987emotion} could be used at any level that is wished upon. If one needs a basic classification, then the ontology could be used up to the primary level, and if one requires their work to have a huge list of emotions, the secondary and tertiary levels of the ontology could be utilized.
% Subsequently, we discovered that 

The organization of the rest of the paper is as follows. Section \ref{related_work} discusses an overview of all the existing ontologies in the emotion domain. We follow that up by discussing various emotion categories in Section \ref{emotion}, focusing primarily on emotion analysis using Parrott's group of emotions. The methodology used for the development of the ontology is detailed in Section \ref{method}, while the proposed ontology is described in Section \ref{proposed_onto}. Section \ref{results} discusses the evaluation along with the results, and Section \ref{applications} depicts the usage of TONE in three areas of computer science. Lastly, Section \ref{future} concludes our findings and probable future works.

\section{Related Work} \label{related_work}
Usage of ontology has multiple perks like generality, ability to reuse, cost-efficiency, and, most importantly, its flexibility. So far, ontologies have been created as a model for some real-life domains and then enhanced incrementally in the future to extend and adapt them to any domain. 

\par Prior works on ontology in the emotion domain date back up to $2005$, when \cite{10.1007/11573548_65} proposed an ontology of emotional cues by grouping the concepts in the ontology to three global modules showing emotion detection or production. At the same time, \cite{10.1007/11573548_45} presented a semantic lexicon of feelings and emotions with an ontology. They also demonstrate how to annotate emotions and use them for textual navigation. In the following year, \cite{GarcaRojas2006EmotionalFE} proposed an ontology to support the modeling of emotional facial animation in virtual humans. Consequently, \cite{GIL2015610} proposed \textit{EmotionsOnto}, which is a generic ontology for describing emotions. They also focus on the detection of emotions and show the efficiency of the ontology by applying it to models that collect emotion common sense.

Alternatively, \cite{sykora2013emotive} targeted the fine-grained detection of emotion in sparse text, like tweets. They introduce an ontology that takes all the basic emotions of $4$ groupings of emotions: Drummond, \cite{Ekman1992-EKMATB}, \cite{izard1971face}, and \cite{plutchik1991emotions}. They remove all those emotions that are less likely on Twitter and hence end up proposing an ontology of $8$ emotions, $8$ basic emotions, negations, intensifiers, conjunctions, interjections, and information on whether individual terms are slang or used in standard English and their associated POS (Parts-of-Speech) tags. In the meantime, for the annotation of emotion in multimedia data, \cite{10.1007/978-3-642-04391-8_32} came up with an ontology to allow an effective sharing of the encoded annotations between different users. in order to manage the emotional content efficiently, \cite{inbook} proposed  \textit{OntoEmotion} with $5$ basic emotions: `Sadness', `Happiness', `Surprise', `Fear', and `Anger'. When the emotional content is expressed as emotional categories, it aims to identify relationships between various levels of specification of related emotions. The work proposed by \cite{tabassum2016emotion} is completely based on building an ontology of emotions based on Plutchik's wheel of emotions \cite{plutchik1991emotions}. Their ontology describes how various colors in the wheel describe the intensities of the emotion and how the entire wheel has helped in creating \textit{EmotiOn} ontology for emotion analysis. 

Alongside emotion-based ontologies, there have also been a few ontologies targeting mental health and various issues related to it. \cite{Hastings2012RepresentingMF} were among the first to introduce an ontology for describing human mental functioning and disease. \textit{Mental Disease Ontology} (MDO) is an ontology that describes and categorizes mental disorders, whereas \textit{Mental Functioning Ontology} (MFO) represents all aspects of mental functioning like cognitive processes, intelligence, etc. In order to model the emotion domain for assisting healthcare professionals in responding to their patient's emotions better, \cite{article} proposed the \textit{Visualized Emotion Ontology} (VEO) to provide a semantic definition of $25$ emotions based on established models, as well as visual representations of emotions utilizing shapes, lines, and colors. All of these ontologies are either task-specific or offer an incomplete ontology that cannot be utilized directly for carrying out all emotion-related tasks. Hence, we introduce an emotion-based ontology, TONE, that addresses various shortcomings of the literature.

\section{Background} \label{emotion}
In order to develop an emotion ontology, available \textbf{models for grouping emotions} could be of assistance and thus in this Section, we elaborate on them. Prior research shows multiple attempts made to classify said emotions by which one may distinguish one emotion from another. \textbf{Dr. William James} \cite{james1950principles}, in $1890$, proposed $4$ basic emotions based on bodily involvement: `Fear', `Grief', `Love', and `Rage'. Contrarily \textbf{Dr. Paul Ekman} \cite{Ekman1992-EKMATB} identified $6$ basic emotions: `Anger', `Disgust', `Fear', `Happiness', `Sadness', and `Surprise'. However, \textbf{Dr. Wallace V. Friesen} and \textbf{Dr. Phoebe C. Ellsworth} worked with Dr. Ekman on the same basic structure and linked emotions to facial expressions. Finally, in the 1990s, Ekman proposed an expanded list of basic emotions, including a range of positive and negative emotions that were not all encoded in facial muscles. The newly included emotions were `Amusement', `Contempt', `Contentment', `Embarrassment', `Excitement', `Guilt', `Pride in achievement', `Relief', `Satisfaction', `Sensory pleasure', and `Shame'. Furthermore, \textbf{Dr. Richard and Dr. Bernice Lazarus} \cite{SLazarus1994-SLAPAR-2}, in $1996$, expanded the list of emotions to $15$ emotions: `Aesthetic experience', `Anger', `Anxiety', `Compassion', `Depression', `Envy', `Fright', `Gratitude', `Guilt', `Happiness', `Hope', `Jealousy', `Love', `Pride', `Relief', `Sadness', and `Shame', in the book \textit{Passion and Reason}\footnote{\url{https://ofoghsalamat.com/ofoghsm/public/files/42A58939-0083-4C20-9071-5AA4AF1CBC32.pdf}}. Researchers at the \textbf{University of California} \cite{cowen2017self}, Berkeley identified $27$ categories of emotion based on $2185$ short videos intended to elicit a certain emotion. In \cite{smith2016book}, \textbf{Dr. Tiffany Watt Smith} listed $154$ different worldwide emotions and feelings. In 1980, \textbf{Dr. Robert Plutchik} diagrammed a wheel of $8$ emotions called \textit{Plutchik's wheel of emotions} \cite{plutchik1991emotions} using `Joy', `Trust', `Fear', `Surprise', `Sadness', `Disgust', `Anger', and `Anticipation'. In $1987$, \textbf{Dr. Gerrod Parrott} \cite{shaver1987emotion} proposed a tree-structured list of emotions centered around $6$ basic emotions: `Anger', `Joy', `Love', `Fear', `Sadness', and `Surprise'. Conclusively, out of all these existing models for the categorization of emotions, \textbf{Parrott's group of emotions} \cite{shaver1987emotion} goes well in creating an ontology as it fits the criteria by being a tree-structured representation \cite{noy2001ontology} and offers the granular distribution of emotions for ease of use. 

\begin{figure}[!t]
    \centering
    \includegraphics[width=0.45\textwidth]{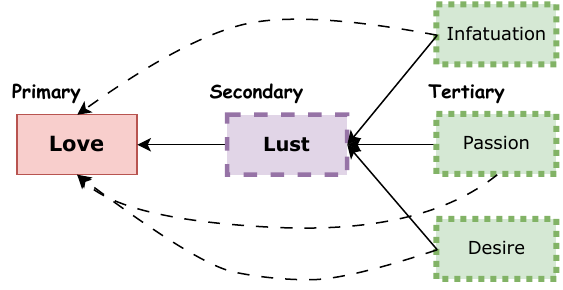}
    \caption{We show the relationship between different tiers i.e., \textit{Primary}, \textit{Secondary}, and \textit{Tertiary} emotions according to Parrott’s group of emotions.}
    \label{fig:relation}
\end{figure}

\begin{figure*}[!t]
    \centering
    \includegraphics[width=\textwidth,height=4.25in]{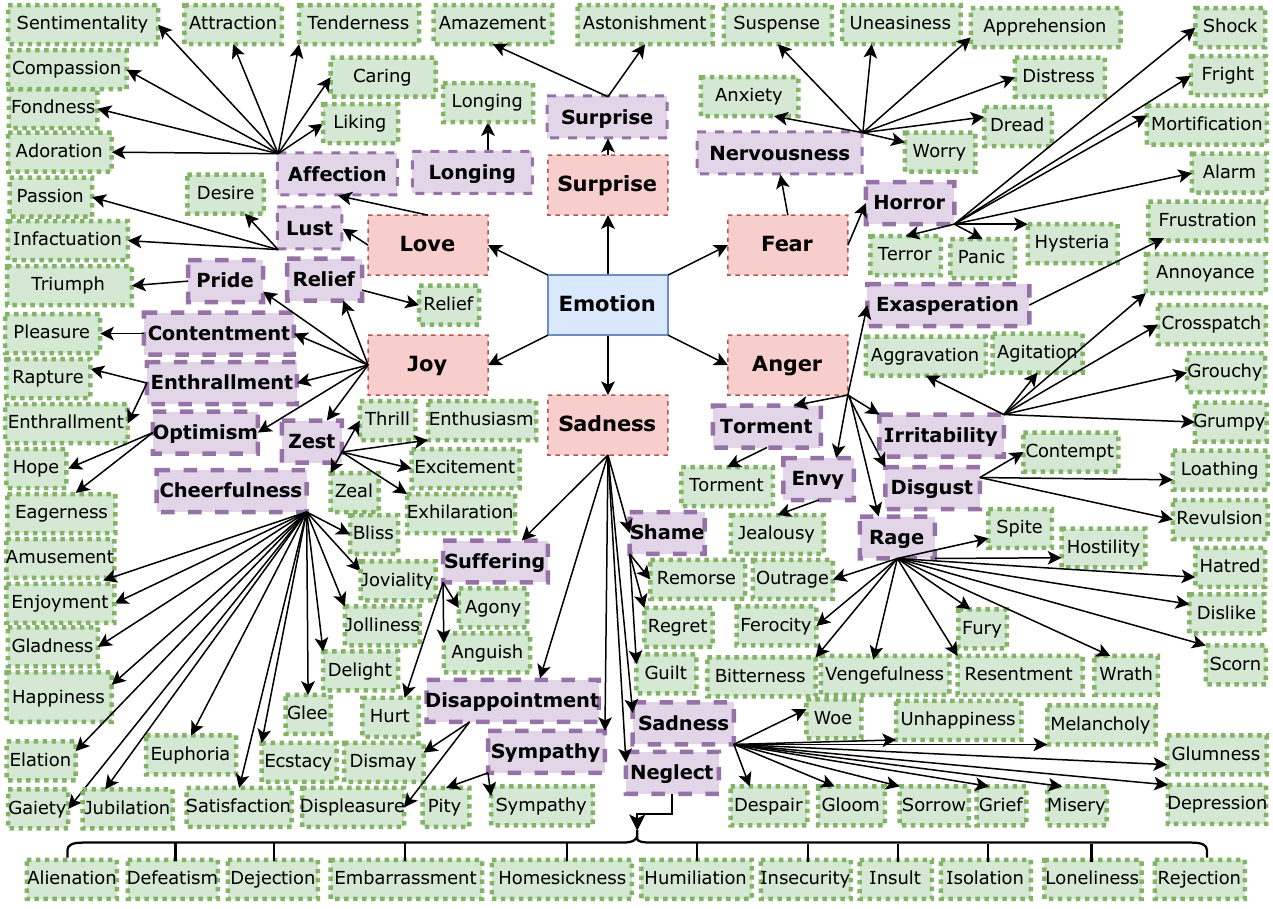}
    \caption{We show Parrott's group of emotions consisting of emotions in $3$ different hierarchies.}
    \label{fig:emotionlist}
\end{figure*}

According to Parrott's group of emotions, there are $3$ tiers depending on their nature and intensities, namely \textit{Primary}, \textit{Secondary}, and \textit{Tertiary}. The entire Parrott's groups of emotions has been represented in Figure~\ref{fig:emotionlist}. These groupings are designed in a way that Primary emotions are the most intense emotions, while tertiary emotions are the least intense emotions. \textit{Primary} emotions, also considered as \textbf{intense} emotions, are represented by \textit{red} boxes showing that those are independent emotions. They are $6$ in number: \textbf{`Anger'}, \textbf{`Fear'}, \textbf{`Joy'}, \textbf{`Love'}, \textbf{`Sadness'}, and \textbf{`Surprise'}. These can be further categorized into numerous sub-categories called the \textit{Secondary} emotions. The \textit{Secondary} emotions are dependent upon primary emotions and are shown by \textit{purple} boxes. They are considered to be \textbf{extreme} emotions. For example, the \textit{Secondary} emotions of `Love' are \textbf{`Affection'}, \textbf{`Lust'}, and \textbf{`Longing'}. Lastly, the \textit{green} boxes represent Tertiary emotions which are said to be directly dependent upon \textit{Secondary} emotions while being indirectly upon \textit{Primary} emotions as shown in Figure~\ref{fig:relation}. For instance, the \textit{Primary} emotion `Love', with \textit{Secondary} emotion `Lust', has \textit{Tertiary} emotions as \textbf{`Desire'}, \textbf{`Passion'}, and \textbf{`Infatuation'}. These \textit{Tertiary} emotions are often referred to as \textbf{mild} emotions. In other words, \textit{Primary} emotions contain their own properties, as well as the properties of their \textit{Secondary} and \textit{Tertiary} emotions. 

\section{Methodology} \label{method}
In this Section, we elaborate on the procedure we follow to build our proposed ontology, TONE: A $3$-\textbf{T}iered \textbf{ON}tology for \textbf{E}motion analysis. Formerly, we mentioned in Section~\ref{introduction} that we utilize \textbf{Parrott's group of emotions} to construct our ontology. We begin with the construction of an extensive vocabulary for each emotion. The next task is to build classes and sub-classes from the hierarchy of the grouping which is followed by different automated methods we use to connect various classes and sub-classes of emotions to each other. This gives us our final ontology and the step-by-step procedure of utilizing emotional groups for ontology creation follows. 

\subsection{Collection of Vocabulary} \label{collecvocab}
Our primary task in the creation of our ontology is to build an extensive vocabulary for each emotion. To do so, we begin by gathering the information needed for the vocabulary and then perform the necessary annotation and verification to finalize it. For each emotion described by Parrott's group of emotions, we must consider two necessary aspects from literature (\cite{10.1007/978-3-319-50112-3_3}, \cite{Liang2006FromAT}, \cite{8665539} etc): the \textbf{definition} of said emotion and its \textbf{synonyms}. For example, if we consider the emotion \textit{`Love'}, we must initially gather its definition and follow this up by gathering its synonyms. This gives us a \textbf{pseudo-vocabulary} of the emotion `Love', which will be refined in future steps. We must repeat the same process of gathering definitions and synonyms for all emotions irrespective of their level. 
% In this instance, we find the definitions of all emotions linked to `Love', like for \textit{Secondary} emotions i.e., \textit{`Affection'}, \textit{`Lust'}, \textit{`Longing'} and all the \textit{Tertiary} emotions. 
In our work, we have used the \textbf{Oxford Dictionary}\footnote{\url{https://www.oxfordlearnersdictionaries.com/}} for gathering the definitions of emotions and \textbf{Thesaurus}\footnote{\url{https://www.thesaurus.com/}} for collecting the synonyms. It is crucial to note that the pseudo-vocabulary contains overlapping synonyms that are stored for the time being. These will be processed in the upcoming steps.

\subsection{Refinement of Vocabulary} \label{refinement}
In this step, we aim to refine the acquired pseudo-vocabulary to obtain the final vocabulary for all the emotions, irrespective of their tier. The previous step gives us a list of different emotions with their definitions and synonyms. This collection of synonyms from Thesaurus is prone to have overlapping words for various emotions. For example, when we consider two \textit{Tertiary} emotions of `Anger' i.e., `Revulsion' and `Loathing', both contain many overlapping synonyms like `Dislike', `Hatred' etc. Our task is to assign these overlapping words to any one emotion as we consider our vocabularies to be mutually exclusive\footnote{In case one wishes to use an overlapping set of vocabulary, they could just extract the synonyms from Thesaurus and start utilizing them as it is by skipping the steps that follow}. In order to assign these overlapping words to any emotion's vocabulary, we use Cosine Similarity between each emotion and every emotion in its pseudo-vocabulary. We use Sentence-BERT\footnote{The presence of phrases in synonyms like `Ill Temper' for the emotion `Anger' necessitates the usage of sentence-level embedding instead of word-level embedding.} \cite{DBLP:journals/corr/abs-1908-10084} to generate representations of the emotions-in-consideration and the overlapping synonym from the pseudo-vocabulary to decide where to place the overlapping word. Sequentially, all overlapping terms are assigned by providing emotions and words as inputs to Sentence-BERT for generating their representations and finding the cosine similarity between them. In our example, to assign `Dislike', we find its similarity score with `Revulsion' and `Loathing'. It turns out that `Dislike' shows a higher similarity score ($0.459>0.415$) with `Loathing' and hence in our pseudo-vocabulary we assign it to `Loathing'.   

To avoid any case of partisanship, we bring in human annotators who could help us verify which overlapping words are to be included under which emotion. In order to select an annotator, we hire $10$ highly educated people with a good grasp of the English language and ask them to take a quiz set up by Britannica\footnote{\url{https://www.britannica.com/quiz/antonyms-and-synonyms}} that tests one's knowledge on synonyms and antonyms. We select the top $4$ individuals with scores above $70\%$ to be our annotators for the task. We take the help of one of them in this step (denoted as $An_1$), and the other three annotators help in the verification of the vocabulary (denoted as $V_1$, $V_2$ and $V_3$) once it is created. 

On finalizing the annotator, his/her task is to assign the overlapping words under the appropriate emotion using the definitions and their own domain knowledge of the topic. We provide $An_1$ with the entire tree structure of emotions, their extracted definitions, and the list of synonyms. His/Her task is to gradually consider each overlapping word, along with the definitions of all emotions, and his/her own comprehension of the words to decide which emotion makes the most sense for the overlapping word. By doing so, we now have two vocabularies, ($1$) the pseudo-vocabulary constructed by us that contains overlapping terms assigned using cosine similarity, and ($2$) the vocabulary created by $An_1$, called \textbf{refined vocabulary}, by assigning overlapping terms using his/her domain knowledge. 

\subsection{Verification of Vocabulary}
In order to meticulously design our ontology, it is fundamental to perform the final sanity check to confirm that the vocabulary is error-free. To ensure this, in this step, we take the help of all three verifiers, $V_1$, $V_2$, and $V_3$, by assigning them the task of comparing both pseudo-vocabulary created by cosine similarity and refined vocabulary created by $An_1$ to finalize where each overlapping word belongs. They are provided with the entire tree structure of emotions, the extracted definitions, the refined vocabulary, and the cosine similarities alongside the pseudo-vocabulary. They must now finalize which synonyms are appropriate to be in which emotion's vocabulary i.e., whether the overlapping words have been duly assigned. In case there is a disagreement regarding the placement of any overlapping word, we take the majority opinion and decisively build our \textbf{final vocabulary}. There was no case of disagreement in our work.

\subsection{Establishing Dependency and Building Final Ontology}
We move on to the creation of the ontology itself which involves the construction of various classes and sub-classes using the groups provided by Dr. Gerrod Parrott and linking the sub-classes to the classes via defined relationships. The construction and declaration of classes and sub-classes using the group of emotions are detailed in Section~\ref{proposed_onto}. Following \cite{uschold1996ontologies}, the construction of classes and sub-classes is followed by linking them together, and there are two directions to do so: ($1$) Linking classes to sub-classes from top to bottom or vice-versa (\textbf{class hierarchy}) and ($2$) Linking classes and sub-classes in a sibling relation (using \textbf{dependencies}). In order to construct the class hierarchy, there are numerous possible ways, and we choose the \textbf{top-down approach} that starts by defining the most general concepts in the domain, following up with subsequent specialization of the concepts. For example, we start with creating classes for the general concepts of Emotion as `Anger', `Fear', `Joy' etc. Then we specialize the `Anger' class by creating some of its sub-classes: `Envy', `Rage', `Disgust' etc. The class hierarchy is shown by using the \textbf{``\textit{is-a}''} axiom \cite{uschold1996ontologies}, which states that ``class A is a sub-class of B if every instance of B is also an instance of A''. In our case, every \textit{Tertiary} emotion \textit{is-a Secondary} emotion and accordingly every \textit{Secondary} emotion \textit{is-a Primary} emotion. The example taken in Figure~\ref{fig:relation} shows, \textbf{`Infatuation' \textit{is-a} `Lust'} and \textbf{`Lust' \textit{is-a} `Love'}. Building such relations became easier because of the hierarchical division of groups of emotions done by Dr. Gerrod Parrott. 

According to \cite{noy2001ontology}, after the creation of classes, sub-classes, necessary vocabulary, and defining the class hierarchy, the final step is to analyze siblings' relations, if any. For that purpose, we use dependencies. \textbf{Dependencies} are properties that help in establishing a relation between any two classes/sub-classes of an ontology. There is no documented method in the literature for creating these dependencies. For that reason, we propose automated methods to establish $3$ types of dependencies in our work.

\begin{figure*}[!t]
\centering
\subfigure[\textit{isOppositeOf}]{
\includegraphics[width=.3\textwidth]{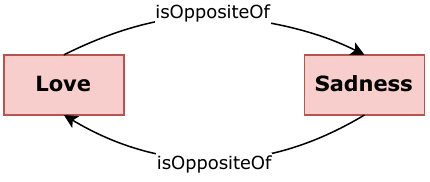}
}
\subfigure[\textit{isComposedOf}]{
\includegraphics[width=.3\textwidth]{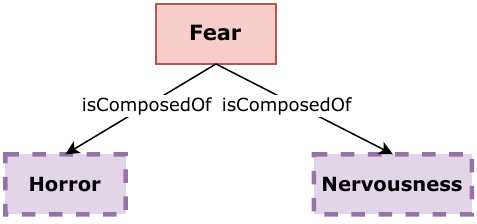}
}
\subfigure[\textit{plus-LeadsTo}]{
\includegraphics[width=.3\textwidth]{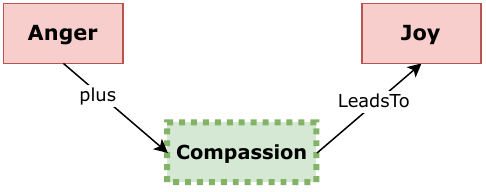}
}
\caption{We show an example for each dependency used in the ontology.}
\label{fig:dependency}
\end{figure*}

\begin{enumerate}
    \item \textbf{\textit{isOppositeOf}}: This dependency connects two contrasting emotions; for example, `Joy' \textit{isOppositeOf} `Sadness'. A noteworthy point is that this dependency can be connected in a $2$-way relationship; for example, if `Joy' \textit{isOppositeOf} `Sadness', then `Sadness' \textit{isOppositeOf} `Joy'. In order to identify the emotions that may be associated with this dependency, we utilize \textit{Thesaurus} to obtain antonyms of each emotion. For example, according to Thesaurus, the antonyms of `Joy' are `Depression', `Melancholy', `Misery', `Sadness', `Sorrow', `Seriousness', `Unhappiness', and a few more. We first check if the considered antonym is a class/sub-class and link them all to `Joy'. Otherwise, we see if it is a part of any emotion's vocabulary. If so, we consider the emotion whose vocabulary contains the word and link that emotion to `Joy'. In our considered example, `Unhappiness' is neither a \textit{Primary}, \textit{Secondary} nor \textit{Tertiary} emotion but is present in the vocabulary of the \textit{Tertiary} emotion `Annoyance'. So we link it as `Joy' \textit{isOppositeOf} `Annoyance'. We automate the aforementioned procedure for connecting the antonyms of each emotion in Algorithm~\ref{alg:opp}. A point to note here is that if we take the antonyms of `Sadness', we may obtain `Joy', and linking these creates a bidirectional relationship as shown in Figure~\ref{fig:dependency}.
    \begin{algorithm}[!h]
    % \scriptsize
    \caption{isOppositeOf()}\label{alg:opp}
    \textbf{Input:} list of all emotions \textbf{$EL$}; \\
    \textbf{Output:} list of opposites of emotions \textbf{$CL$}; \\
    $CL \xleftarrow{} [];$\\
    \ForEach{$\text{emotion } i \in EL$}{
    $antonyms_i \xleftarrow{} \text{Retrieve all antonyms of } i \text{ from Thesaurus};$\\
    % $ CL_i.append(antonyms_i);$\\
    \For{$j \in antonyms_i$}{\eIf{$j \text{ is present in EL and} j \neq i$}{
    $antonyms_i.keep(j);$\\
    }{
    \If{$j \text{ is present in vocabulary of any emotion } e \in EL$}{$antonyms_i.replace(j, e);$\\}
    % $antonyms_i.discard(j);$\\
    }}
    $CL.append(i, antonyms_i);$\\}
    \end{algorithm}
    \item \textbf{\textit{isComposedOf}}: This dependency connects all the sub-classes to the superclass; for example, Figure~\ref{fig:dependency} shows an example of how `Fear' \textit{isComposedOf} `Nervousness' and `Horror'. The tiered structure of Parrott's group of emotions comes of immense help while assigning these dependencies. We introduce this dependency to guarantee that emotions can be utilized at any level, based on the granularity of one's task, without the fear of missing out on lower-level vocabularies. So in case one wishes to put only \textit{Primary} emotions to use, they could use the \textit{isComposedOf} dependency to converge the vocabulary of the sub-classes of each emotion to enrich the vocabulary of all the \textit{Primary} emotions. In order to accurately establish links between emotions using this dependency, we define an automated mechanism in Algorithm~\ref{alg:comp}. 
    \begin{algorithm}[!h]
    \caption{isComposedOf()}\label{alg:comp}
    \textbf{Input:} the list of all emotions $EL$, the list of \textit{Primary} emotions \textbf{$PEL$}, the list of \textit{Secondary} emotions \textbf{$SEL$}, the list of \textit{Tertiary} emotions \textbf{$TEL$}, and tree structure $T$; \\
    \textbf{Output:} list of set of emotions $SL$; \\
    $SL \xleftarrow{} [];$\\
    \ForEach{$\text{emotion } i \in EL$}{
    \uIf{$\text{emotion } i \in PEL$}{
    \ForAll{$\text{emotion } j \in SEL$}{
    % \eIf{$i \text{ is T.parent and } j \text{ is T.child of } i$}{
    \If{$j \text{ is child of } i \text{ in } T$}{
    $SL.append(\{j,i\});$\\
    }
    % {$discard(\{j,i\});$\\}
    }}
    \uElseIf{$\text{emotion } i \in SEL$}{
    \ForAll{$\text{emotion } j \in TEL$}{
    % \eIf{$i \text{ is T.parent and } j \text{ is T.child of } i$}{
    \If{$j \text{ is child of } i \text{ in } T$}{
    $SL.append(\{j,i\});$\\
    }
    % {$discard(\{j,i\});$\\}
    }}
    \Else(\tcc*[f]{$\text{Since } i \in TEL, T.child == None$})
    {$continue;$}
    }
    \end{algorithm}
    \item \textbf{\textit{plus-LeadsTo}}: This dependency connects three emotions by taking into account the possibility that any combination of any two emotions might lead to the third. In our case, we check if any \textit{Primary} emotion \textit{plus} any \textit{Secondary} or \textit{Tertiary} emotion \textit{LeadsTo} another \textit{Primary} emotion. Due to a lack of data for emotion detection, we limit our study to only \textit{Primary} emotion in the case of \textit{LeadsTo}. For example, Figure~\ref{fig:dependency} shows how the \textit{Primary} emotion `Anger', in combination with a \textit{Tertiary} emotion `Compassion' helps in obtaining the \textit{Primary} emotion of `Joy'.

    In order to automate this dependency, we perform the steps mentioned in Algorithm~\ref{alg:leads}. Firstly we automatically generate a short sentence portraying each \textit{Primary}, \textit{Secondary}, and \textit{Tertiary} emotion. For example, `Joy' is depicted using \textit{`I am very happy'} or `Fear' can be depicted using \textit{`I am very scared'}. For each emotion, we formulate such a statement. Secondly, we repeatedly consider each \textit{Primary} emotion and concatenate it with all \textit{Secondary} and \textit{Tertiary} emotions to create an integrated sentence.  For example, statements of `Anger' and `Guilt' result in \textit{`I am very scared. I feel very guilty'}. Thirdly, we run an emotion classifier \footnote{We train a RoBERTa classifier on the Kaggle dataset: \textbf{Emotions dataset for NLP} \url{https://www.kaggle.com/praveengovi/emotions-dataset-for-nlp}} to detect the emotion of the integrated sentence and our considered example detects the resultant emotion as `Sadness'. Finally, the dependency links are established for all the emotions as in our example it would be: `Anger' \textit{plus} `Guilt' \textit{LeadsTo} `Sadness'. We perform a second-level verification by asking verifiers $V_1$, $V_2$ and $V_3$ to go through these automatically created links to obtain a flawless ontology. 
    \begin{algorithm}[!h]
    \caption{plus-LeadsTo()}\label{alg:leads}
    \textbf{Input:} list of all emotions $EL$, the list of \textit{Primary} emotions \textbf{$PEL$}, the list of automatically generated statements for each emotion $Stmt$ ; \\
    \textbf{Output:} list of set of plus emotions $PL$, list of set of LeadsTo emotions $LL$ ; \\
    $Pl \xleftarrow{} [];$\\
    $LL \xleftarrow{} [];$\\
    $combstring \xleftarrow{} \text{` '};$\\
    $emodetect \xleftarrow{} \text{` '};$\\
    \ForEach{$\text{emotion } i \in PEL$}{
    \ForEach{$\text{emotion } j \in EL$}{
    \If{$j \notin PEL$}{
    $combstring = Stmt(i) \oplus Stmt(j);$\\
    $emodetect \xleftarrow{} EmotionClassification(combstring);$\\
    }
    \If{$emodetect \neq i$}{
        $PL.append(\{i,j\});$\\
        $LL.append(\{j,emodetect\});$
    }
    }
    }
    \end{algorithm}
\end{enumerate}

We use these dependencies to define a relationship between distinct emotions in our created tiers to generate our final ontology, TONE. 
% But we wish to verify if our designed ontology is precise. To do so, we bring our final verifier, $V_2$, to go through all the available materials like definitions, similarity scores, etc., to make a decision on whether the created ontology is credible. 

\section{TONE: The Proposed Ontology} \label{proposed_onto}
\begin{figure*}
    \centering
    \includegraphics[width=\textwidth]{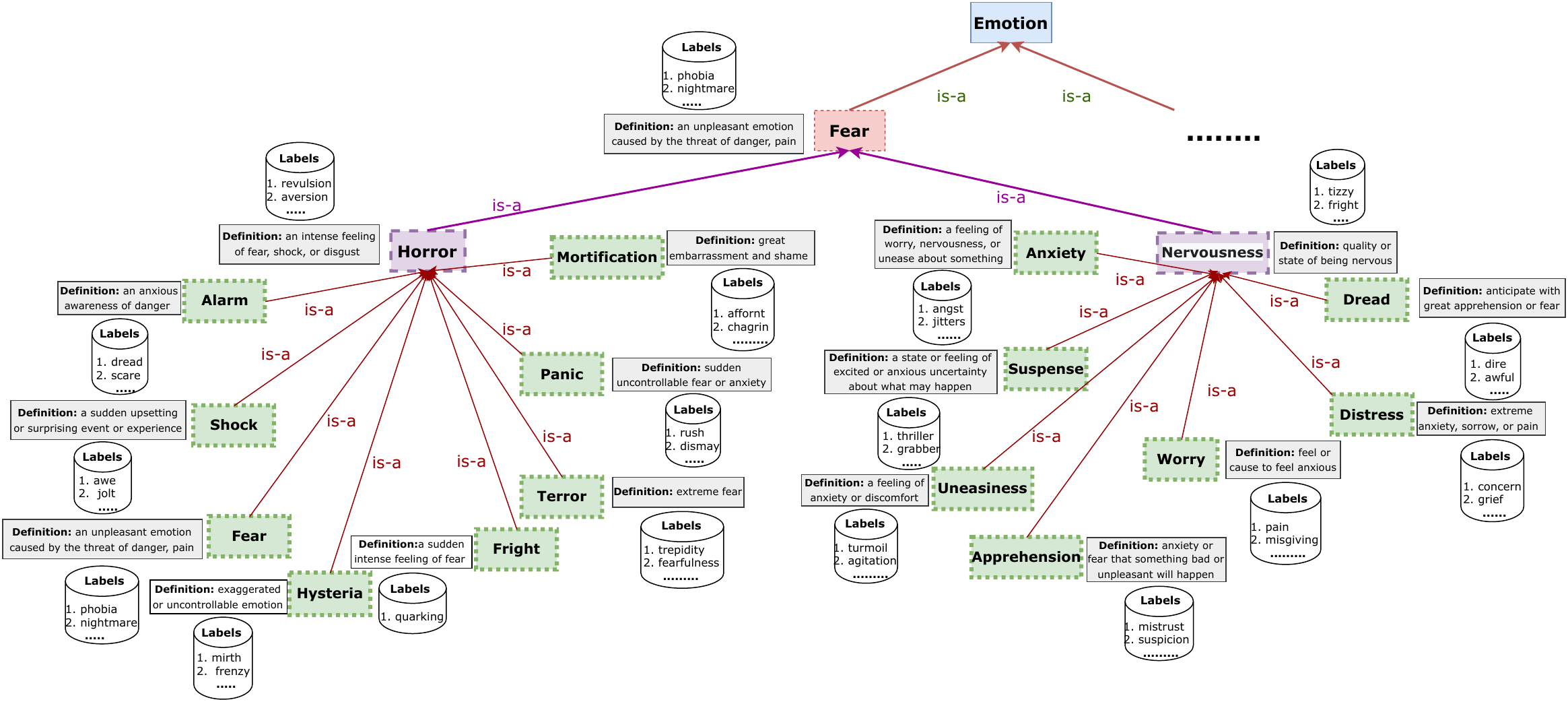}
    \caption{We show a snippet of the ontology with the \textit{Primary} emotion as \textbf{`Fear'}, its \textit{Secondary} emotions: \textbf{`Horror'}, \textbf{`Nervousness'} and all their \textit{Tertiary} emotions alongside their respective definitions and vocabularies.}
    \label{fig:ontosnap}
\end{figure*}

Our proposed ontology is entirely based on Parrott's groups of emotions and is called TONE: A $3$-\textbf{T}iered \textbf{ON}tology for \textbf{E}motion analysis\footnote{This ontology is available at \url{https://github.com/srishtigupta253/TONE.git}.}, as shown in Figure~\ref{fig:ontosnap}. In this Section, we define and declare all the classes, sub-classes, and the object properties we use in the creation of the ontology. We follow the existing literature (\cite{tabassum2016emotion}, \cite{8665539}) and use the tool, \textbf{Protege}\footnote{\url{https://protege.stanford.edu/}} to generate our ontology.

% \subsection{Class}
\textbf{Classes} in an ontology describe concepts in the domain. A class can have multiple \textbf{sub-classes} that represent concepts that are more granular than the superclass. In our case, \textbf{Emotion} is the root of the hierarchy which has $6$ classes pertaining to the $6$ \textit{Primary} emotions that exist in the top-most level of the hierarchy. Each class has multiple sub-classes like for the ontology root, Emotion, `Fear' is one of the $6$ classes, and its sub-classes are `Horror' and `Nervousness', as shown in Figure~\ref{fig:ontosnap}.

We use the Web Ontology Language (\textbf{OWL})\footnote{\url{https://www.w3.org/TR/owl2-overview/}} as the ontology language for representing TONE. In order to illustrate the classes, OWL uses the `$<$Declaration$>$ $</$Declaration$>$' element to declare each class and allows prefixing the \textbf{IRI} (Internationalized Resource Identifier) of the ontology. IRI is an identifier that uniquely recognizes each element of an ontology. For any IRI of an ontology, any entity is identified by its name followed by the character ‘\#’. Prefixing helps to use the IRI subsequently throughout the ontology for adding new entities without having to re-specify it each time. A class is declared as follows:
\begin{verbatim}
    <Declaration>
        <Class IRI="#Fear"/>
    </Declaration>
    <Declaration>
        <Class IRI="#Nervousness"/>
    </Declaration>
\end{verbatim}

Note that these are just the declarations. Any other information or relation, such as sub-classes and disjoint classes, etc., are specified separately within their own respective elements. Hence, initially, all emotions are declared as classes, following which each emotion is linked to its superclass as in this example below:
\begin{verbatim}
    <SubClassOf>
        <Class IRI="#Nervousness"/>
        <Class IRI="#Fear"/>
    </SubClassOf>
    <SubClassOf>
        <Class IRI="#Uneasiness"/>
        <Class IRI="#Nervousness"/>
    </SubClassOf>
\end{verbatim}

Each sub-class is defined under $<$SubClassOf$>$ element. The first $<$Class$>$ element identifies the sub-class, and the second $<$Class$>$ element is the parent/superclass. Here, we can see that the \textit{Tertiary} emotions are represented as sub-classes of the \textit{Secondary} emotions, and similarly to show the tier structure, \textit{Primary} emotions become the superclass of \textit{Secondary} emotions. Hence, the representations remain the same, but further in OWL, properties are defined to differentiate them tier-wise. Similarly, we use $<$DisjointClasses$>$ element in OWL to represent the classes that are distinct from each other (disjoint classes) as: 
\begin{verbatim}
    <DisjointClasses>        
        <Class IRI="#Joy"/>       
        <Class IRI="#Sadness"/> 
    </DisjointClasses> 
\end{verbatim}

In Section \ref{collecvocab}, we mention that definitions for each emotion word are collected, and apart from the vocabulary, every emotion at each tier has these definitions. The Protege tool lets us reflect these definitions in the ontology as follows:
\begin{verbatim}
    <AnnotationAssertion>
        <AnnotationProperty IRI="#definition"/>
        <IRI>#Nervousness</IRI>
        <Literal datatypeIRI="&xsd;string">the
        quality or state of being nervous</Literal>
    </AnnotationAssertion>    
\end{verbatim}

As mentioned earlier, we wish to build a vocabulary for all emotions at every tier, and these built vocabularies must be disclosed in the ontology. These built vocabularies are linked to the corresponding emotion and represented as:
\begin{verbatim}
    <AnnotationAssertion>
        <AnnotationProperty abbreviatedIRI=
                "rdfs:label"/>
        <IRI>#Nervousness</IRI>
        <Literal datatypeIRI="&rdfs;Literal">
                touchiness</Literal>
    </AnnotationAssertion>
    <AnnotationAssertion>
        <AnnotationProperty abbreviatedIRI=
                "rdfs:label"/>
        <IRI>#Nervousness</IRI>
        <Literal datatypeIRI="&rdfs;Literal">
                trembles</Literal>
    </AnnotationAssertion>
    <AnnotationAssertion>
        <AnnotationProperty abbreviatedIRI=
                "rdfs:label"/>
        <IRI>#Nervousness</IRI>
        <Literal datatypeIRI="&rdfs;Literal">
                tremulousness</Literal>
    </AnnotationAssertion>
    <AnnotationAssertion>
        <AnnotationProperty abbreviatedIRI=
                "rdfs:label"/>
        <IRI>#Nervousness</IRI>
        <Literal datatypeIRI="&rdfs;Literal">
                willies</Literal>
    </AnnotationAssertion>
\end{verbatim}

Various dependencies have been used in our work, and the following are their declarations. For each dependency, we use a built-in property that links a class description to another class description called $<$EquivalentClasses$>$. We define the \textbf{\textit{plus-LeadsTo}} dependency below to show the examples considered in Section~\ref{method}. 
\begin{verbatim}
    <EquivalentClasses>
        <Class IRI="#Compassion"/>
        <ObjectExactCardinality cardinality="1">
            <ObjectProperty IRI="#Plus"/>
            <Class IRI="#Anger"/>
        </ObjectExactCardinality>
    </EquivalentClasses>
    
    <EquivalentClasses>
        <Class IRI="#Joy"/>
        <ObjectExactCardinality cardinality="1">
            <ObjectProperty IRI="#LeadsTo"/>
            <Class IRI="#Compassion"/>
        </ObjectExactCardinality>
    </EquivalentClasses>
\end{verbatim}

Similarly, for the \textbf{\textit{isComposedOf}} dependency, we define the domain of the dependency as the superclass and the range of the dependency as the sub-class. An instance of representing this using $<$EquivalentClasses$>$ is shown below.
\begin{verbatim}
   <EquivalentClasses>
        <Class IRI="#Fear"/>
        <ObjectSomeValuesFrom>
            <ObjectProperty IRI="#isComposedOf"/>
            <Class IRI="#Horror"/>
        </ObjectSomeValuesFrom>
    </EquivalentClasses>
    
    <EquivalentClasses>
        <Class IRI="#Fear"/>
        <ObjectSomeValuesFrom>
            <ObjectProperty IRI="#isComposedOf"/>
            <Class IRI="#Nervousness"/>
        </ObjectSomeValuesFrom>
    </EquivalentClasses>
\end{verbatim}

Finally, we show the \textbf{\textit{isOppositeOf}} dependency, where we define that `Joy' \textit{isOppositeOf} `Sadness'. In this case, the domain is `Joy', and the range would be `Sadness', but, as mentioned in Section~\ref{method}, the domain and range can be interchangeable if a cycle exists.
\begin{verbatim}
    <EquivalentClasses>
        <Class IRI="#Sadness"/>
        <ObjectSomeValuesFrom>
            <ObjectProperty IRI="#isOppositeOf"/>
            <Class IRI="#Joy"/>
        </ObjectSomeValuesFrom>
    </EquivalentClasses>
\end{verbatim}
\section{Evaluation and Results} \label{results}
In this Section, we show the type of tests we perform in order to test various aspects of TONE and their corresponding results. As shown by \cite{gomez2001evaluation}, the lack of metrics for the evaluation of ontologies makes it difficult to verify them. For that reason, we use two well-known types of evaluation procedures: ($1$) \textbf{Human Evaluation}, where the annotators manually verify for various things of our ontology like correctness of hierarchy, classes, dependencies, etc., and ($2$) \textbf{Automatic Evaluation}, where we automatically evaluate TONE by running DL (Description Logic) queries over the HermiT reasoner\footnote{\url{http://www.hermit-reasoner.com/}} to check the consistency of the ontology. The procedure for both the evaluations and their corresponding results are mentioned next.

\subsection{Human Evaluation}
We analyze the quality of TONE by performing human evaluations on various aspects. For this evaluation, we asked $10$ Ph.D. scholars (called \textit{authenticators} from here on) with a background in either emotions or ontology to rate our ontology on various fronts in the range of $1-5$ with $1$ being the worst rating and $5$ being the best. In this evaluation, we perform a qualitative analysis of our ontology using $2$ different aspects: ($1$) Structural evaluation and ($2$) Semantic relation evaluation, whose details follow. 

\subsubsection{Structural Evaluation}
To perform the structural evaluation, we follow \cite{8665539} and evaluate TONE on the basis of the following facets: ($1$) The \textbf{correctness} of the hierarchical structure: assessed on the basis of the `\textit{is-a}' relationship as to whether the given class A is in the \textit{is-a} relationship with class B. ($2$) The \textbf{expressiveness}: check the efficacy of the ontology to identify relevant information. ($3$) The \textbf{completeness} of the ontology: evaluate whether the assigned definitions and vocabulary are able to define various emotional scenarios. For instance, let us consider the \textit{Tertiary} emotion of `Love', i.e., `Caring'. One must verify the definition of the emotion (`displaying kindness and concern for others' in our example of `Caring') and the labels assigned as its vocabulary (`Friendly', `Loving', `Sympathetic', `Warm', `Attached' etc. of `Caring') for the completeness of the ontology. ($4$) The \textbf{clarity} of the ontology: the class/sub-class name in the ontology must be meaningful and easy to understand. Moreover, one could also perform a \textbf{redundancy} check to check whether the vocabulary has any duplicates, which is verified in Section~\ref{method} with the help of the annotator ($An_1$) and verifiers ($V_1$, $V_2$ and $V_3$).
\begin{figure}[!h]
    \centering
    \includegraphics[width=0.475\textwidth]{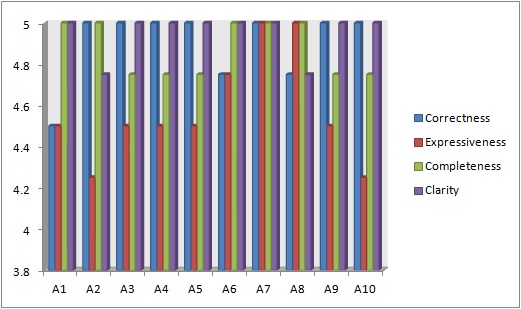}
    \caption{We show the results obtained after the Structural Evaluation of our ontology, TONE.}
    \label{fig:structural}
\end{figure}
% 	Correctness	Expressiveness	Completeness	Clarity
% A1	4.5	4.5	5	5
% A2	5	4.25	5	4.75
% A3	5	4.5	4.75	5
% A4	5	4.5	4.75	5
% A5	5	4.5	4.75	5
% A6	4.75	4.75	5	5
% A7	5	5	5	5
% A8	4.75	5	5	4.75
% A9	5	4.5	4.75	5
% A10	5	4.25	4.75	5

Figure~\ref{fig:structural} shows the results obtained from $10$ authenticators on all four fronts. TONE achieves great results in all four areas implying that the ontology is efficient and satisfyingly associated in terms of classes and sub-classes. To summarize the results, we obtain an average rating of $4.9$ for its correctness, $4.57$ for its expressiveness, $4.87$ for its completeness, and $4.95$ for its clarity.

\begin{figure}[!h]
    \centering
    \includegraphics[width=0.45\textwidth]{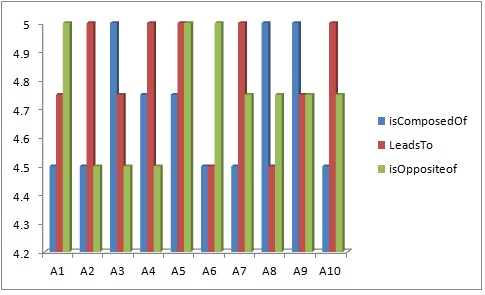}
    \caption{We show the results obtained after the Semantic Evaluation of our ontology, TONE.}
    \label{fig:semantic}
\end{figure}

\subsubsection{Semantic Relation Evaluation}
In an ontology, one of the most salient aspects is dependencies, as they are responsible for logically linking the classes together. In this evaluation, we ask the authenticators to go through the ontology entirely to judge the three considered dependencies in our work, as mentioned in Section~\ref{method}. The task of the authenticators is to rate the assignment of each dependency assigned by our automated methods to link the hierarchy of emotions. We depict the results obtained in Figure~\ref{fig:semantic}, and it shows that most of the authenticators agree that the dependencies have been appropriately placed. The authenticators, whose ratings fell short of $5$ were asked to give feedback regarding our scope of improvement. Their suggestions were appropriately incorporated and re-verified by the rest of the authenticators to reach a common ground and finalize our ontology. These results hence demonstrate that TONE is semantically well-related and shows a clear link between various emotions. 
% 	isComposedOf	LeadsTo	isOppositeof
% A1	4.5	4.75	5
% A2	4.5	5	4.5
% A3	5	4.75	4.5
% A4	4.75	5	4.5
% A5	4.75	5	5
% A6	4.5	4.5	5
% A7	4.5	5	4.75
% A8	5	4.5	4.75
% A9	5	4.75	4.75
% A10	4.5	5	4.75

\subsection{Automatic Evaluation}
In order to perform an automatic evaluation of our ontology, we formulate various queries and feed them to the \textbf{HermiT reasoner} as DL (Description Logic) queries. These queries are simple forms of expression to check: ($1$) if the ontology has appropriately defined the domain, classes, and its properties and ($2$) if the classes are appropriately linked together or not.
\begin{figure*}[!t]
\centering
\subfigure[\textit{isOppositeOf}]{
\includegraphics[width=.22\textwidth,height=1.45in]{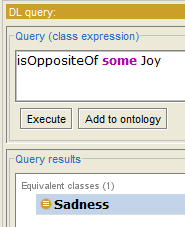}
}
\subfigure[\textit{isComposedOf}]{
\includegraphics[width=.22\textwidth,height=1.45in]{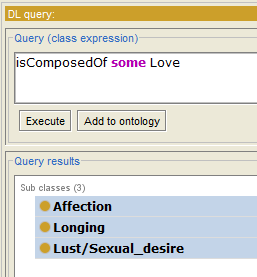}
}
\subfigure[\textit{plus-LeadsTo}]{
\includegraphics[width=.33\textwidth,height=1.45in]{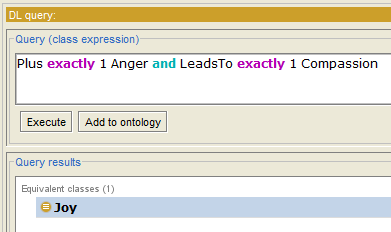}
}
\caption{We show an example of different outputs obtained by running various DL queries on HermiT Reasoner.}
\label{fig:queries}
\end{figure*}

We formulate several queries to test our ontology upon, and Figure~\ref{fig:queries} shows one such query with its result. The purpose of this query was to retrieve all the contradictory emotions of `Joy'. The result shows that `Sadness' was appropriately returned as our dependencies labeled the relation between them via the \textit{`isOppositeOf'} property. Similarly, we run a query to retrieve all the sub-classes that `Love' is made up of. In case the ontology had to be used at the \textit{Primary} emotion level, \textit{isComposedOf} dependency can be run as shown in Figure~\ref{fig:queries}, and the results are accurately displayed. Lastly, we check if the \textit{plus-LeadsTo} dependency helps in understanding the product obtained by combining $2$ emotions. We run the example shown in Figure~\ref{fig:dependency}, and Figure~\ref{fig:queries} shows that we successfully obtained the result for the query. Similarly, we run queries to check each relationship and hence conclude that our ontology is appropriately bound after performing both human and automatic evaluations.

\section{Applications} \label{applications}
In this Section, we demonstrate how our proposed ontology, TONE, can be used in highly emergent fields of computer science that utilize emotions. We show the influence of TONE in $3$ different domains: (Section~\ref{ed}) \textbf{Emotion Detection}, (Section~\ref{review}) \textbf{Product Reviews} and (Section~\ref{erg}) \textbf{Empathetic Response Generation}.  
% In this Section, we elaborate on how TONE can be used in $2$ . The proposed ontology might be useful in the field of emotion in a variety of ways, and we present $4$ ways of utilizing TONE in the cause extraction task and generating empathetic responses. 

\subsection{Emotion Detection} \label{ed}
The first use-case that we demonstrate is the area of emotion detection which analyzes various sentences to predict which emotion it portrays. In this experiment, we wish to portray how our extended vocabulary at each tier of the hierarchy could be immensely beneficial. We consider $6$ emotions that are the \textit{Primary} emotions in our ontology. The \textit{isComposedOf} dependency comes to use here for enriching the vocabulary of each \textit{Primary} emotion combining their corresponding \textit{Secondary} and \textit{Tertiary} emotions. For instance, `Fear' is made up of `Horror' and `Nervousness'; thus, we must combine the \textit{Tertiary} emotion vocabulary and the individual vocabulary of \textit{Secondary} emotions `Horror' and `Nervousness' to create the final vocabulary for `Fear'. 

Subsequently, we compare emotion detected using our proposed vocabulary with three different works: ($1$) SCAECE proposed by \cite{gupta2023scaece}, ($2$) Tom Drummond's proposed vocabulary \footnote{\url{https://tomdrummond.com/leading-and-caring-for-children/emotion-vocabulary/}} of emotions and ($3$) Geneva Emotion Wheel proposed by \cite{doi:10.1177/0539018405058216}. \cite{gupta2023scaece} perform the detection task using the same $6$ \textit{Primary} emotions in Parrott's group of emotions. They create a vocabulary manually to check the presence of an emotional word in various utterances of a conversation. We confirmed with them that they created a vocabulary of $755$ instances for the $6$ \textit{Primary} emotions and acquired it for comparison. Secondly, we consider Tom Drummond's proposed vocabulary of emotions, which consists of $474$ instances of emotion words spread across 10 emotions. Similarly, the Geneva Affect Label Coder has $279$ instances of emotions across 20 emotion categories. We ask the same annotator $An_1$ (from Section~\ref{refinement}) to help us re-categorize these emotion vocabularies into required $6$ \textit{Primary} emotions for ease of comparison.

\begin{table}[!h]
    \caption{We show the comparison of TONE with various vocabularies of emotion to prove its efficiency in detecting emotion of $100$ test sentences.}
    \centering
    \begin{tabular}{|c|c|}
       \hline
       \textbf{Model} & \textbf{Detected Sentences}\\ \hline
       SCAECE & 92\\ 
       Drummond & 82\\
       GALC & 75 \\
       TONE & \textbf{97} \\ \hline
    \end{tabular}
    \label{tab:emodet}
\end{table}

To carry out the task of comparison, we gather $100$ test sentences from \textit{Random Sentence Generator} \footnote{\url{https://randomwordgenerator.com/sentence.php}} and test the efficiency of each vocabulary for the detection task. In order to compare the results, we ask annotator $An_1$ from Section~\ref{refinement} to help us generate the ground truth of the considered $100$ test sentences. From the findings of Table~\ref{tab:emodet}, we conclude that the vocabulary used by \cite{gupta2023scaece} is able to detect the emotion $92$ times out of $100$ sentences, Drummond's vocabulary succeeds only 82 times, and the Geneva Affect Label Coder results in $75$ only correct predictions. Contrastively, our proposed vocabulary detects the emotion word in \textbf{$97$} sentences. This could be because, in our introduced vocabulary, we have nearly $1500$ words combining all $6$ emotions' vocabularies, which is nearly double the vocabulary of \cite{gupta2023scaece} and triple of Drummond's. Hence, we show how our introduced ontology can help escalate the Emotion Detection task. 

\subsection{Prediction of Helpful Reviews} \label{review}
\cite{Martin_Pu_2014} works on first checking if the presence of emotions in the text is enough to decide if a given review is part of the helpful ones or not. Following this, they also confirm if the presence of emotions allows an accurate estimation of the number of users who found this particular review helpful. The comparison shown by \cite{Martin_Pu_2014} utilizes $3$ datasets among which the only publicly available dataset is \textbf{Yelp reviews}\footnote{\url{http://www.kaggle.com/c/yelp-recsys-2013}} and hence we show the improvement TONE can bring on it. In order to predict the presence of emotion, they selected Scherer’s emotion model \cite{doi:10.1177/0539018405058216} (called Geneva Emotion Wheel) with its $20$ emotion categories and then used the Geneva Affect Label Coder (GALC) lexicon \cite{doi:10.1177/0539018405058216} to determine which words were conveying which emotion. They compare their proposed model by extracting feature vectors with the number of occurrences of each emotion based on GALC. They run an SVM algorithm on this extracted feature. We perform the same task but use TONE instead of GALC for feature extraction and test the effect TONE can cause against the same experiments. Our comparison considers two versions of our ontology, ($1$) TONE, which involves all $144$ emotions and their vocabularies which produces a feature vector of $144$ features. ($2$) P-TONE, which produces a feature vector of $6$ features by involving only \textit{Primary} emotions and vocabularies of each emotion obtained by combining its corresponding \textit{Secondary} and \textit{Tertiary} emotions. 

\begin{table}[!t]
    \centering
    \caption{We show the evaluation of the SVM algorithm on the features extracted over Yelp datasets using Accuracy (A), F1-score (F1), and Area Under the Curve (AUC) as measures.}
    \begin{tabular}{|c|c|c|c|}
    \hline
    \textbf{Model} & \textbf{A} & \textbf{F1} & \textbf{AUC}\\ \hline
    RATE & 0.50 & 0.60 & 0.52\\ 
    LEN1 & 0.73 & 0.75 & 0.80\\ 
    LEN2 & 0.72 & 0.73 & 0.79\\ 
    POS & 0.84 & 0.87 & 0.85\\ 
    SENTI & 0.63 & 0.63 & 0.67\\ 
    FLES & 0.53 & 0.61 & 0.53\\ 
    GEWN & 0.65 & 0.68 & 0.70\\ 
    GEW20 & 0.65 & 0.68 & 0.69\\ 
    GEW21 & 0.73 & 0.75 & 0.80\\ 
    P-TONE & 0.63 & 0.66 & 0.68\\
    TONE & \textbf{0.85} & \textbf{0.88} & \textbf{0.89}\\ 
    \hline
    \end{tabular}
    \label{tab:prediction}
\end{table}

From Table~\ref{tab:prediction}, we can clearly see that by evaluating the SVM algorithm by using emotion features extracted using TONE, there is a significant improvement in the accuracy of predicting helpful reviews. This could be because of the enhanced vocabulary provided by our ontology. In addition, if the number of emotions for comparison increases, as in the case of P-Tone, the model becomes less efficient. Hence, we show the use case of two variants of TONE in this line of work.

\subsection{Empathetic Response Generation} \label{erg}
Finally, we also depict how multiple dependencies of TONE could be used in the Empathetic Response Generation task. The main agenda here of responding with empathy is that one must focus on being positive and redirect all of the speaker's negative emotions with empathy. To achieve this, it is vital to know what the target positive emotion is when one notices a negative emotion in the utterances of an ongoing conversation. The \textit{isOppositeOf} dependency becomes a huge aid in this case as it could help in determining which emotion the conversational agent should react with. Once the dialog agent finds out which emotion to react with using the \textit{isOppositeOf} property, the agent must also know how to achieve it. The \textit{plus-LeadsTo} dependency shows the path to the dialog agent for the possibility of swaying a negative emotion to a positive emotion. For instance, if the speaker expresses `Anger', we could use `Compassion' because the \textit{plus-LeadsTo} dependency depicted in Figure~\ref{fig:dependency} contributes to the production of a positive feeling. Thus, by indicating which emotion to add, this dependency may aid in turning negative feelings into pleasant ones. In such a task, the dependencies that \textit{LeadsTo} negative emotions can be ignored as empathetic responses would not need the paths that lead to a pessimistic emotion like `Anger'.
\begin{figure}[!t]
    \centering
    \includegraphics[width=0.45\textwidth]{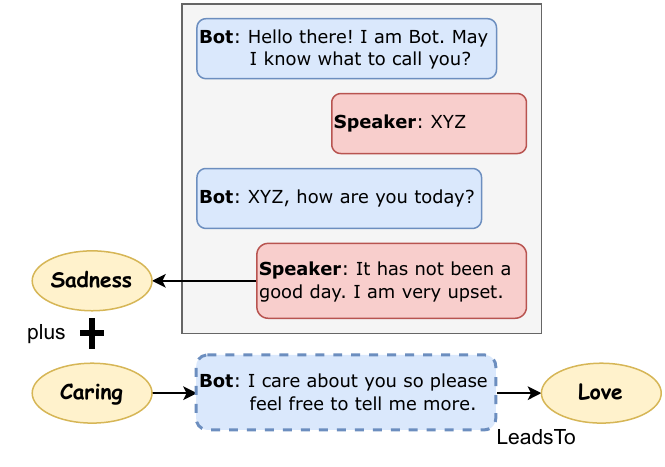}
    \caption{We show an example of how TONE can help improve empathetic responses in a dialog agent.}
    \label{fig:erg}
\end{figure}

\section{Conclusion and Future Work} \label{future}
TONE was created to guide all emotion categorization and relationship-related tasks. In our work, we showed how Parrott's group of emotions may be utilized to construct an ontology based on the hierarchy of emotions. At each tier, we also establish a suitable vocabulary for each emotion using our proposed semi-automated mechanism. Furthermore, we propose automated approaches for creating $3$ distinct sorts of dependencies to link emotions at different tiers of the hierarchy. Finally, the efficacy of the ontology is evaluated using both automated and human evaluation. The results of the human evaluation show that our ontology is widely preferred by annotators. In this work, we also show the usage of various aspects of this ontology in computer science applications like Empathetic Response Generation and Cause Extraction tasks. After performing the necessary experiments, we observe that TONE helps in improving the results of the Emotion Detection task by $5-22\%$ and the Accuracy of Product Review Prediction by $7-31\%$. The ontology has been created using semantic web language (OWL). We aim to extend the ontology in the following directions: ($1$) attempt to enhance existing emotion-related Natural Language Processing tasks like emotion classification using this ontology and ($2$) work on automating the enrichment of vocabulary as currently the vocabulary is static and needs human effort. 

\printcredits

\bibliographystyle{cas-model2-names}
\bibliography{cas-refs}
\newpage
\bio{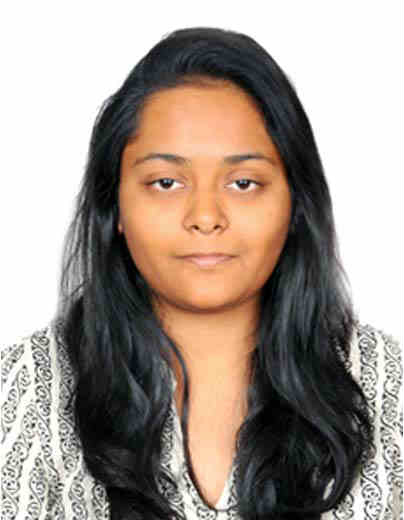}
\textbf{Srishti Gupta} : is a PhD Scholar with the Department of Computer Science and Engineering, Indian Institute of Technology, Patna, India. She received her M.Tech. degree from Jawaharlal Nehru Technological University Hyderabad, India, in 2020, and B.Tech degree from Jawaharlal Nehru Technological University Hyderabad, India, in 2018. Her current research interests include Dialogue Agents, Empathetic Response, and Natural Language Processing.
\endbio
\bio{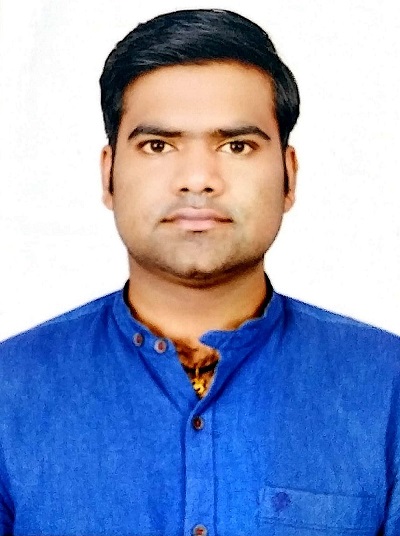}
\textbf{Piyush Kumar Garg} : is a PhD Scholar with the Department of Computer Science and Engineering, Indian Institute of Technology, Patna, India. He received the M.Tech. degree from IIT(ISM) Dhanbad, India in 2018 and B.Tech degree from the College of Technology and Engineering, Udaipur, India in 2015. His current research interests include social network analysis, crisis response, and information retrieval.
\endbio
\bio{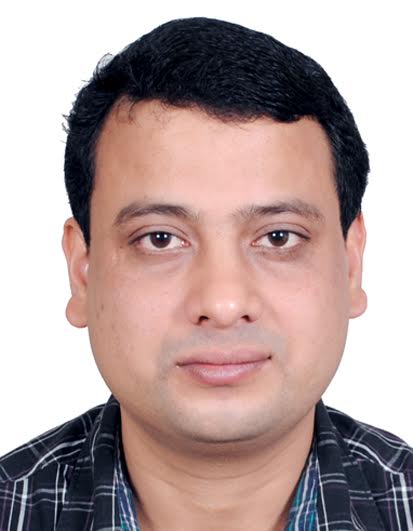}
\textbf{Sourav Kumar Dandapat} : is an Assistant Professor at the Indian Institute of Technology Patna from 2016, February onward. He completed his PhD in 2015 and M.Tech in 2005 from the Indian Institute of Technology Kharagpur, India. He received his B.E degree from Jadavpur University, West Bengal, India in 2002. His current research interest includes Computational Journalism, Social Computing, Information Retrieval, Human-Computer Interaction, etc.
\endbio

\end{document}